%% file: main.tex
\documentclass{article}

\usepackage[dvipsnames]{xcolor}
\usepackage{microtype}
\usepackage{graphicx}
\usepackage{booktabs} 
\usepackage{listings}
\usepackage{hyperref}

\usepackage[accepted]{icml2024}

\usepackage{amsmath}
\usepackage{amssymb}
\usepackage{mathtools}
\usepackage{amsthm}
\usepackage{comment}
\usepackage{caption}
\usepackage{subcaption}

\usepackage[capitalize,noabbrev]{cleveref}
\usepackage{caption}
\usepackage{graphicx}
\usepackage{framed}
\usepackage{multirow}
\theoremstyle{plain}

\theoremstyle{definition}

\theoremstyle{remark}

\usepackage[textsize=tiny]{todonotes}

\icmltitlerunning{AnyTool: Self-Reflective, Hierarchical Agents for Large-Scale API Calls}

\begin{document}

\twocolumn[
\icmltitle{AnyTool: Self-Reflective, Hierarchical Agents for Large-Scale API Calls}

\icmlsetsymbol{equal}{*}

\begin{icmlauthorlist}
\textbf{Yu Du}\textsuperscript{\rm 1 *}\quad\quad
\textbf{Fangyun Wei}\textsuperscript{\rm 2 * $\dagger$}\quad\quad
\textbf{Hongyang Zhang}\textsuperscript{\rm 3} \\
\textsuperscript{\rm 1}Tsinghua University \quad\quad
\textsuperscript{\rm 2}Microsoft Research Asia \quad\quad
\textsuperscript{\rm 3}University of Waterloo \\
\texttt{ duyu20@mails.tsinghua.edu.cn}\quad
\texttt{fawe@microsoft.com}\quad
\texttt{hongyang.zhang@uwaterloo.ca} \\
\textsuperscript{*} Equal contribution \quad\quad \textsuperscript{$\dagger$}~\text{Corresponding author}\\
\end{icmlauthorlist}

\vskip 0.3in
]

\input{sections/0-abstract}
\input{sections/1-introduction}

\input{sections/2-related_work}
\input{sections/3-method}

\input{sections/4-experiments}

\input{sections/5-conclusion}
\input{sections/7-impact}

\bibliography{references}
\bibliographystyle{icml2024}
\input{sections/6-appendix}

\end{document}

%% file: sections/0-abstract.tex
\begin{abstract}
We introduce AnyTool, a large language model agent designed to revolutionize the utilization of a vast array of tools in addressing user queries. We utilize over 16,000 APIs from Rapid API, operating under the assumption that a subset of these APIs could potentially resolve the queries. AnyTool primarily incorporates three elements: an API retriever with a \emph{hierarchical} structure, a solver aimed at resolving user queries using a selected set of API candidates, and a \emph{self-reflection} mechanism, which re-activates AnyTool if the initial solution proves impracticable. AnyTool is powered by the function calling feature of GPT-4, eliminating the need for training external modules. We also revisit the evaluation protocol introduced by previous works and identify a limitation in this protocol that leads to an artificially high pass rate. By revising the evaluation protocol to better reflect practical application scenarios, we introduce an additional benchmark, termed AnyToolBench. Experiments across various datasets demonstrate the superiority of our AnyTool over strong baselines such as ToolLLM and a GPT-4 variant tailored for tool utilization. For instance, AnyTool outperforms ToolLLM by +35.4\% in terms of average pass rate on ToolBench. Code will be available at \href{https://github.com/dyabel/AnyTool}{https://github.com/dyabel/AnyTool}.

\end{abstract}

%% file: sections/1-introduction.tex
\vspace{-8mm}
\section{Introduction}

\begin{figure}[!t]
     \centering
     \begin{subfigure}[!t]{0.95\linewidth}
         \centering
\includegraphics[width=0.95\linewidth]{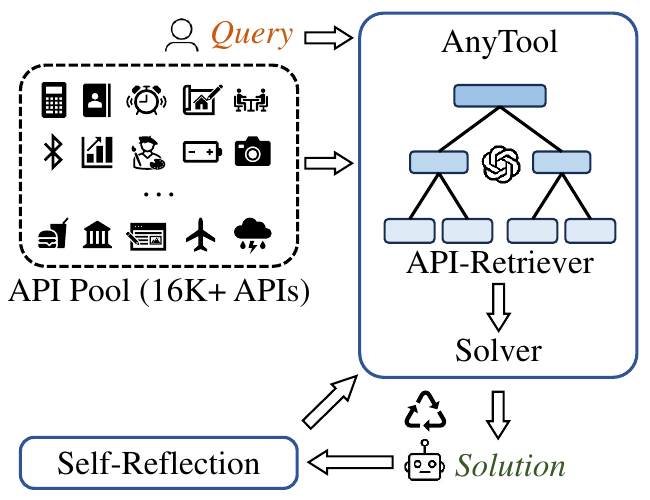}
\vspace{-1mm}
         \caption{AnyTool addresses user queries by leveraging 16k+ APIs. It integrates a hierarchical API-retriever, a solver, and a self-reflection mechanism in a closed loop, all operating without the need for additional training.}
         \label{fig:teaser-1}
     \end{subfigure}
     \hfill
     \begin{subfigure}[!t]{0.99\linewidth}
         \centering
         \includegraphics[width=0.99\linewidth]{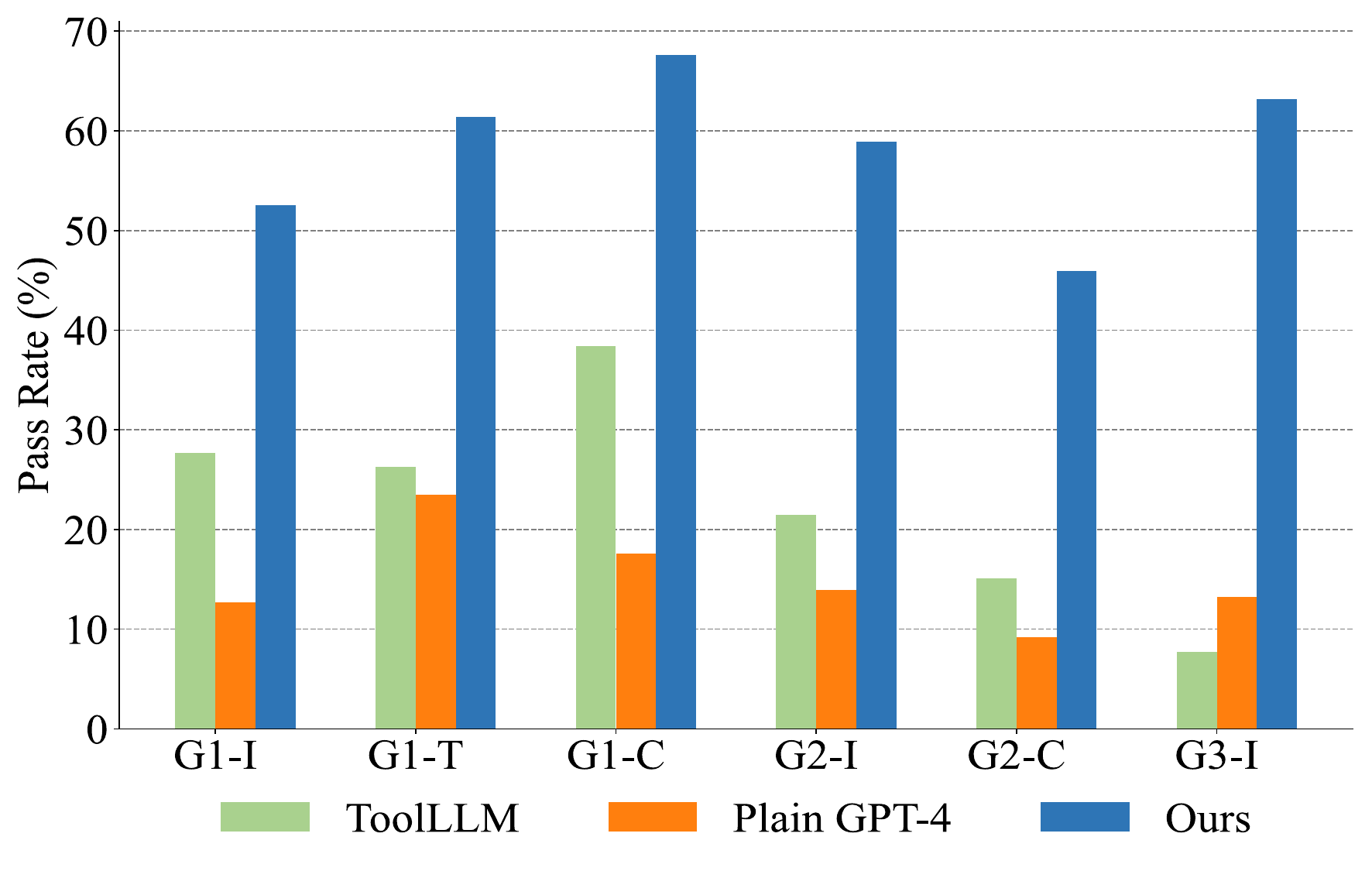}
         \vspace{-2mm}
         \caption{Comparison with ToolLLM and a GPT-4 variant tailored for tool utilization across six subsets of ToolBench~\cite{qin2023toolllm}, using pass rate defined in Eq \ref{eq:pass_rate} as the evaluation metric.}
         \label{fig:teaser-2}
     \end{subfigure}
     \vspace{-2mm}
     \caption{(a) Illustration of AnyTool. (b) Comparison in performance.}
     \vspace{-5mm}
\end{figure}

\begin{figure*}[!t]
    \centering
\includegraphics[width=0.99\linewidth]{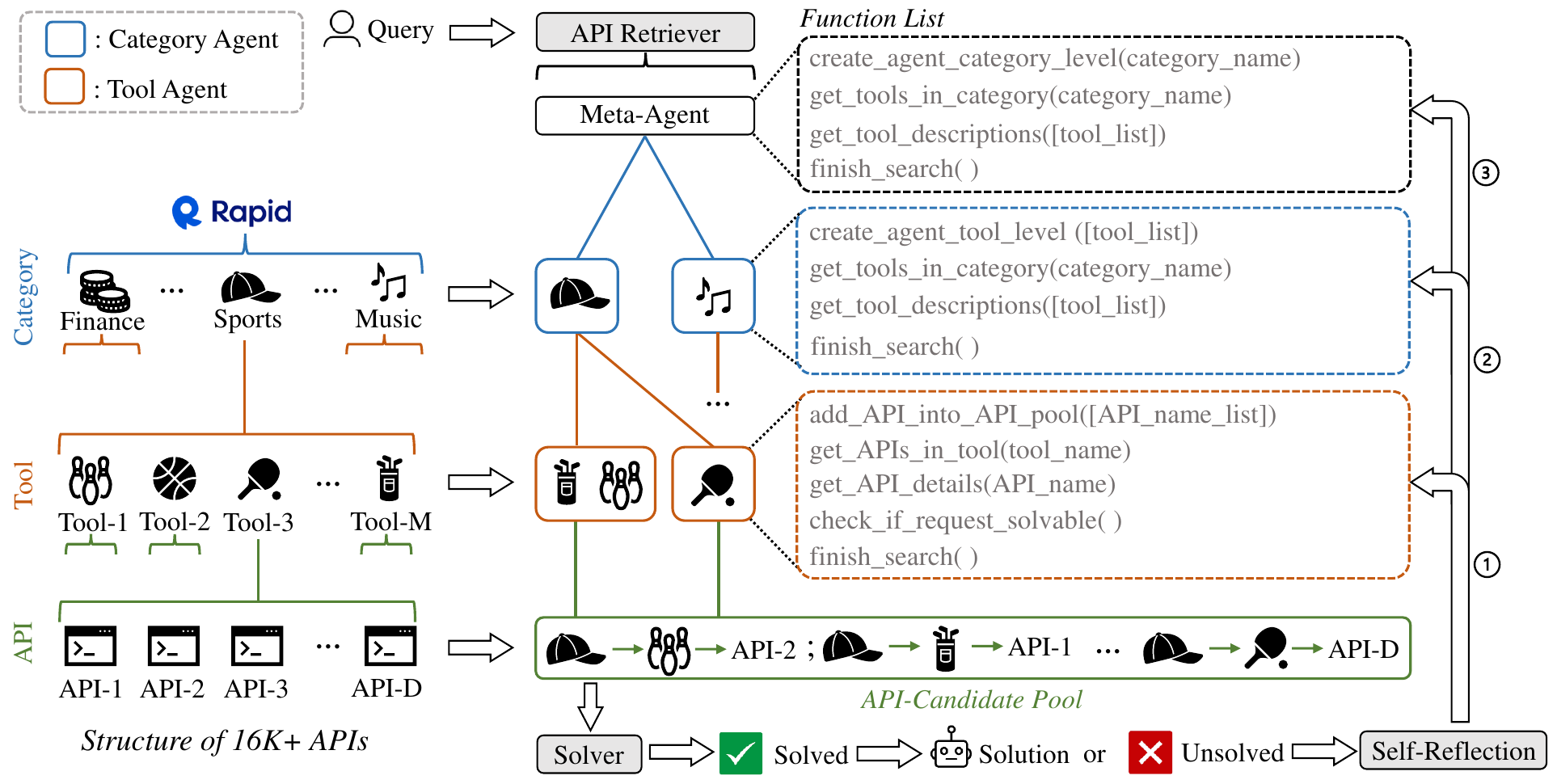}
\vspace{-3mm}
    \caption{Overview of AnyTool. It primarily consists of a hierarchical API retriever tasked with identifying the most relevant API candidates to the user query from a large API pool, a solver aimed at addressing the queries using the generated \textcolor{OliveGreen}{API-candidate pool}, and a self-reflection mechanism. The hierarchical structure includes a \textit{meta-agent} linked with several \textcolor{RoyalBlue}{category agents}, each of which manages a collection of \textcolor{Bittersweet}{tool agents}. We leverage the API structure defined by Rapid API as a guideline. Each type of agent is assigned several functions that it can use to explore the API space. Refer to Table~\ref{tab:agent_function_list} in the appendix for the details of each function.}
    \vspace{-3mm}
    \label{fig:method}
\end{figure*}

From the dawn of civilization, humanity has embarked on a relentless journey of discovery and innovation, mastering an ever-expanding array of tools to enhance our capabilities and increase production efficiency. As we have evolved, so have our tools, transitioning from simple stone implements to complex machines and beyond. Today, we stand at the forefront of a new era, reaping the benefits of the rapid developments in artificial intelligence, particularly the recent advances in large language models (LLMs)~\cite{gpt3,touvron2023llama,touvron2023llama2,chowdhery2023palm,achiam2023gpt,ouyang2022training}. A pivotal challenge now is learning how to drive LLMs to effectively use tools~\cite{qin2023tool,xu2023tool,cai2023large,song2306restgpt,ruan2023tptu,shen2023hugginggpt,hao2023toolkengpt}, a task that could redefine our interaction with technology. Towards this end, we introduce AnyTool, a GPT-4-empowered agent, as depicted in Figure~\ref{fig:teaser-1}. It is designed to effectively leverage more than 16,000 APIs to address user queries, with a significant performance leap as depicted in Figure~\ref{fig:teaser-2}.

\vspace{-1mm}
Previous research~\cite{qin2023toolllm} formulated tool utilization in a dual-phase approach: initially retrieving, then resolving. Specifically, the first phase involves retrieving the most pertinent APIs from a substantial collection of 16K+ APIs in response to user queries. The subsequent phase focuses on utilizing these chosen APIs to address user queries. Our AnyTool uses this design principle while introducing four distinct characteristics (see Figure \ref{fig:method} for an overview):

\vspace{-1mm}
\textit{Plug-and-Play.} Our AnyTool does not require the training of any modules, except for the function-calling feature of GPT-4~\cite{achiam2023gpt}. This aspect sets it apart from existing methods like ToolLLM, which necessitates training an API retriever capable of selecting a set of candidate APIs from the API pool~\cite{qin2023toolllm}.

\vspace{-1mm}
\textit{Hierarchical Structure.} To identify the most relevant APIs for user queries, we design a hierarchical structure within our API retriever. This structure is composed of three tiers, each containing one or multiple agents with diverse roles. This arrangement is inspired by the divide-and-conquer approach. Additionally, we effectively incorporate the API categorization suggested by Rapid API into our hierarchical structure. Consequently, this significantly reduces the search scope for each agent and overcomes constraints related to the maximum context length in LLMs.

\begin{figure}[t!]
    \centering
\vspace{-8mm}
\includegraphics[width=0.99\linewidth]{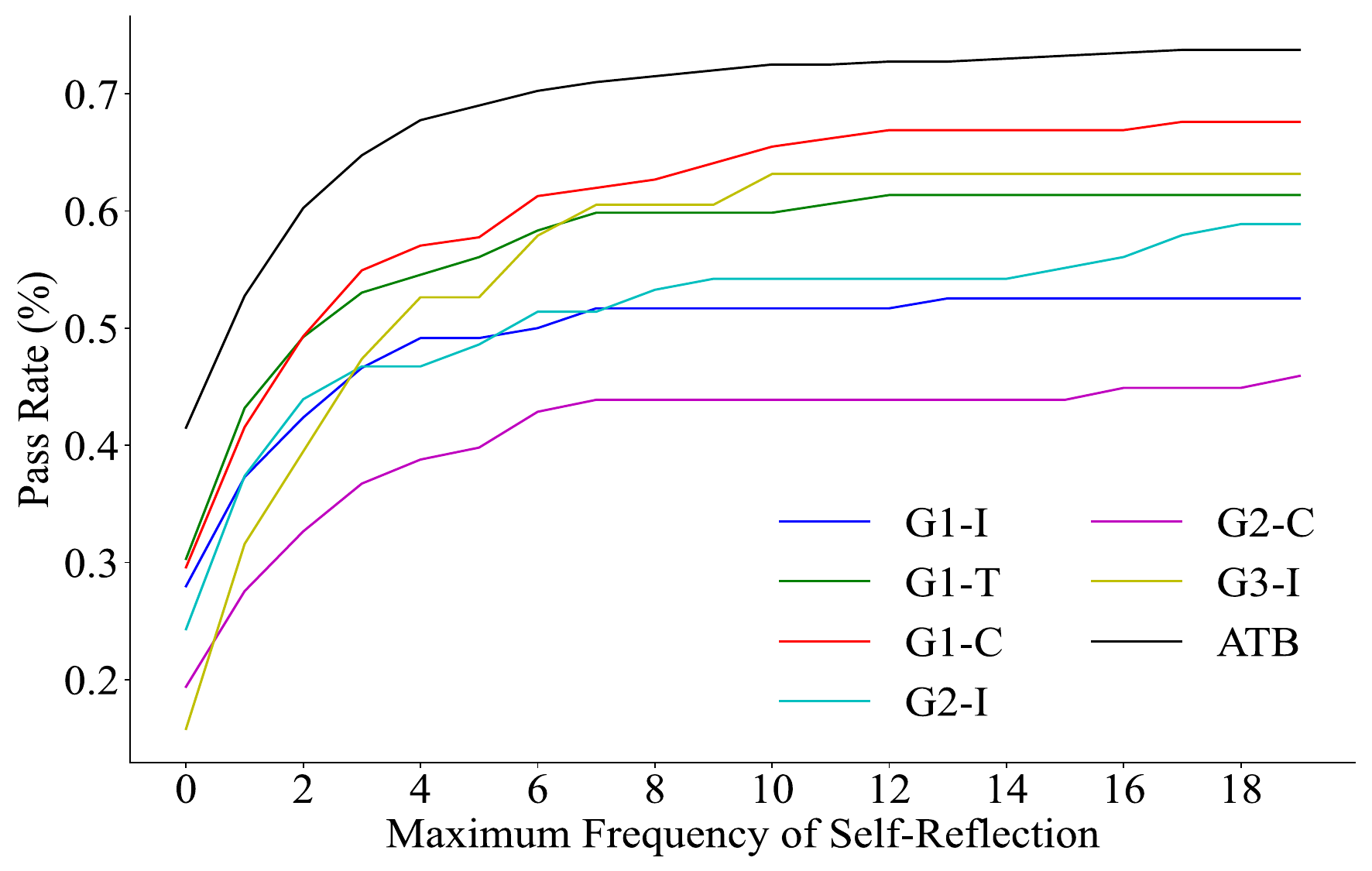}
\vspace{-3mm}
    \caption{The performance of our AnyTool on different datasets (each denoted by a curve) improves as the number of self-reflection rounds increases. ATB: AnyToolBench.}
    \vspace{-5mm}
\label{fig:pass_rate_freq_vs_sr_freq}
\end{figure}

\vspace{-1mm}
\textit{Self-Reflection Mechanism.} Our AnyTool is designed to address user queries through a process of initial attempt followed by reflection. Upon receiving a query, AnyTool suggests a solution, which is then evaluated for feasibility by GPT-4. In cases where the proposed solution is deemed impractical, AnyTool is re-activated, with the consideration of reasons for failure and relevant historical contexts. This mechanism significantly reduces the tendency to ``over-search'' for simpler queries, while also providing a more context-rich and in-depth search for complex queries. This closed-loop system enhances the efficiency and effectiveness of the query resolution process. Figure~\ref{fig:pass_rate_freq_vs_sr_freq} shows how the pass rate improves w.r.t. the self-reflection rounds. With only 4-6 self-reflection iterations, the pass rate improves by up to 20\% across all datasets.

\vspace{-1mm}
\textit{Evaluation for Realistic Scenarios.} The evaluation framework presented in ToolBench~\cite{qin2023toolllm} commences with categorizing user queries as either solvable or non-solvable, employing a set of reference APIs. Following this, the solvable queries undergo further scrutiny to determine if they are successfully addressed or not. However, for those non-solvable queries, the evaluation system regards them as solved when calculating the pass rate, leading to an artificially high pass rate. Our study delves into the intricacies of this evaluation methodology and proposes a revised protocol that better mirrors practical application scenarios.

\vspace{-1mm}
In addition to evaluation on ToolBench, we introduce an extra benchmark, termed AnyToolBench, to facilitate the application of our new evaluation protocol. Experimentally, AnyTool achieves state-of-the-art performance, surpassing strong baselines such as ToolLLM and a version of GPT-4 specifically tailored for tool utilization across various datasets, as illustrated in Figure~\ref{fig:teaser-2}.

%% file: sections/2-related_work.tex
\vspace{-2mm}
\section{Related Works}

\textbf{Tool Utilization in LLMs.}
Large language models~\cite{gpt1,gpt2,gpt3,touvron2023llama,touvron2023llama2,thoppilan2022lamda} may commit factual errors when responding to queries, particularly struggling with precise numbers and specific fields of expertise~\cite{huang2023large,augenstein2023factuality}. Utilizing tools can help mitigate this issue~\cite{li2023api,qin2023toolllm, parisi2022talm,tang2023toolalpaca,hsieh2023tool,schick2023toolformer}. Previous work has involved using an API retriever to match relevant APIs from a large API pool based on the documents, employing either an pretrained text embedding model~\cite{li2023api,patil2023gorilla} or one finetuned with curated API retrieval data~\cite{qin2023toolllm}. However, this approach typically suffers from low accuracy and may overlook the truly relevant APIs. Moreover, there is a lack of feedback mechanism in their retrieval, often leading to unsolved queries due to incorrect API candidates being provided. Our AnyTool fills this gap by directly using the GPT-4 as the API retriever with a hierarchical structure design, and introduces the self-reflection mechanism into the whole process.

\vspace{-1mm}
\textbf{Self-Reflection Mechanism in LLMs.} Self-reflection is a featured ability of LLMs. It was first studied in the LLM alignment problems. \citet{wang2022self} considered the ability of GPT-3 to self-generate instructions for alignment finetuning. Without finetuning, \citet{li2024rain} introduced an inference method, RAIN, that allows pre-trained LLMs to evaluate their own generation and use the evaluation results to guide rewind and generation for AI safety. Recently, \citet{chen2024self} proposed a self-play mechanism, where the LLM refines its capability by playing against instances of itself. \citet{yuan2024self} proposed self-rewarding language models, where the language model itself is used via LLM-as-a-Judge prompting to provide its own rewards for the following DPO finetuning~\cite{rafailov2023direct}. On the other hand, some negative results on self-reflection were also investigated. For example, \citet{huang2023large} showed that GPT-3.5-Turbo and GPT-4 cannot self-correct reasoning yet. But whether GPT-4 can serve as a self-reflective agent for API calling remains an open problem in the existing literature.

%% file: sections/3-method.tex
\vspace{-2mm}
\section{Preliminaries}
\vspace{-1mm}
\begin{figure*}[!t]
    \centering
\includegraphics[width=0.99\linewidth]{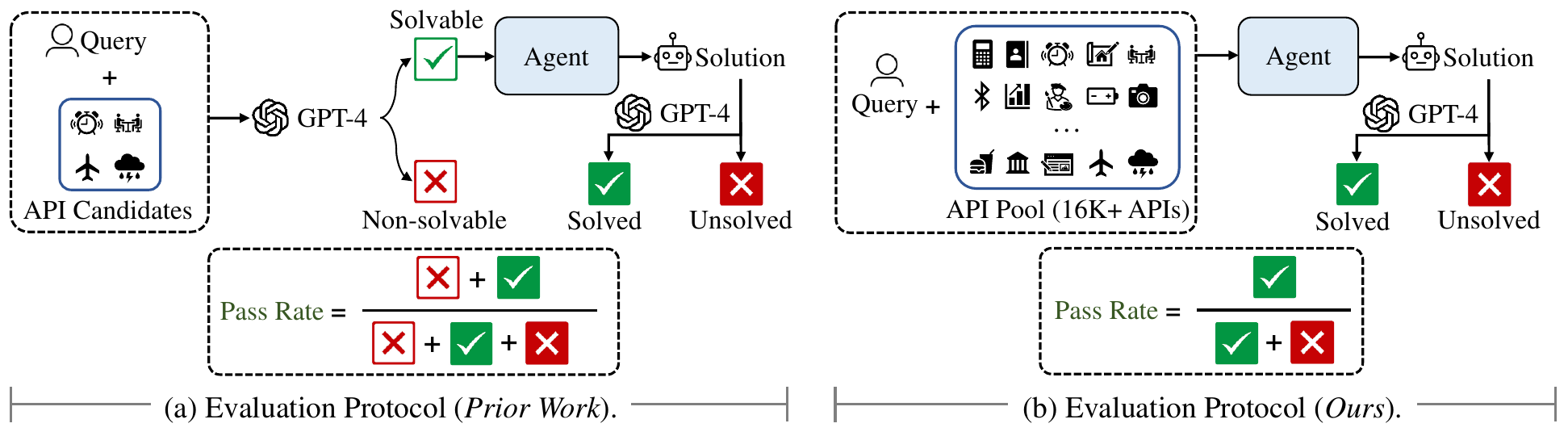}
\vspace{-2mm}
    \caption{Illustration of the evaluation protocols used by: (a) ToolLLM~\cite{qin2023toolllm}; and (b) ours. In (a), if the API retriever selects candidates completely unrelated to the user’s query, GPT-4 may classify all queries as “non-solvable”, leading to an artificially high pass rate, despite the queries remaining unsolved. In (b), we conduct a manual review of all queries and retain only those queries that can be resolved with specific APIs from the API pool for ToolBench.}
    \vspace{-2mm}
    \label{fig:overview}
\end{figure*}
\subsection{Function Calling} 
\vspace{-1mm}
\label{sec:function_calling}
Function calling is a core characteristic of GPT-4~\cite{achiam2023gpt}. Specifically, in response to a user's query $\mathcal{Q}$, the function calling system accesses a set of $M$ distinct functions $\{\mathcal{F}_i\}_{i=1}^{M}$. Each function $\mathcal{F}_i$ has the potential to solve $\mathcal{Q}$, a part of $\mathcal{Q}$, or may not be relevant to $\mathcal{Q}$ at all. The functionality of $\mathcal{F}_i$ is elaborated in a specific document that outlines its purpose, required and optional parameters along with their explanations, the types of output it generates, and the interpretations of these outputs. Note that the function calling feature of GPT-4 does not require visibility into the detailed implementations of each function. It understands their intentions and functionalities through linguistic comprehension.

The process of function calling involves: 1) the user inputs both the query $\mathcal{Q}$ and the function list $\{\mathcal{F}_i\}_{i=1}^{M}$, alongside a designated ``Finish Function'' $\mathcal{F}^*$, into GPT-4; 2) GPT-4 generates a function calling request for the user, with clear input parameters; 3) the user executes the specific function and provides the historical context and function response to GPT-4; 4) this cycle of steps two and three is repeated multiple times until GPT-4 activates the ``Finish Function'' $\mathcal{F}^*$, signaling the resolution of query $\mathcal{Q}$. Users have the option to either employ the output of $\mathcal{F}^*$ directly, or to gather the interim results generated during the function calling process, according to their specific goals or design.

\vspace{-2mm}
\subsection{Problem Formulation and Evaluation}
\vspace{-1mm}
\label{sec:problem_evaluation}
\textbf{Problem Formulation.} The objective of this work is to develop a proficient agent capable of utilizing a vast collection of real-world APIs to address user queries. We use over 16K real-world APIs from the RapidAPI Hub, as collected in the ToolLLM~\cite{qin2023toolllm}. These APIs are represented as $\{\text{API}_{i}\}_{i=1}^{N}$, forming our API pool. The effectiveness of the solutions generated by the agent is assessed using GPT-4. This evaluation involves processing both the user query $\mathcal{Q}$ and the proposed solution $\mathcal{S}$, in accordance with established evaluation protocols and criteria, to ascertain the solution's ability to adequately address the query. \textit{We have also conducted human evaluation and find a correlation as high as 96.5\% between GPT-4 and human evaluations.}

\vspace{-1mm}
\textbf{Evaluation Protocol.} We first revisit the evaluation protocol initially introduced by ToolLLM~\cite{qin2023toolllm}. ToolLLM employs a dual-phase approach for utilizing various APIs. In the first phase, an API retriever is developed to select the most relevant API candidates from the API pool according to a user query $\mathcal{Q}$. The second phase involves ToolLLaMA, a specialized agent that formulates a solution using the selected API candidates. Due to its dual-phase nature, ToolLLM's evaluation is twofold. Initially, GPT-4 evaluates whether the selected API candidates can address the query $\mathcal{Q}$, categorizing them as either ``solvable'' or ``non-solvable''. If a query is deemed ``solvable'', GPT-4 then assesses the effectiveness of the provided solution, classifying it as either ``solved'' or ``unsolved''. Figure~\ref{fig:overview}(a) illustrates how the pass rate $R$ is calculated:
\vspace{-1mm}
\begin{equation}
\label{eq:toolllm_evaluation}
    R = \frac{\textbf{\#}(\text{Non-solvable}) + \textbf{\#}(\text{Solved})}{\textbf{\#}(\text{Non-solvable}) + \textbf{\#}(\text{Solved}) + \textbf{\#}(\text{Unsolved})}.
\end{equation}
However, a significant flaw exists in this evaluation protocol. If the API retriever selects candidates completely unrelated to the user's query, GPT-4 may classify all queries as ``non-solvable'', leading to an artificially high pass rate, despite the queries remaining unsolved. Our experimental evidence confirms this issue, showing that when API candidates are randomly selected for each query, GPT-4 predominantly labels them as ``non-solvable'', resulting in an inflated pass rate of 99.0\% through the metric defined in Eq~\ref{eq:toolllm_evaluation}.

To address the limitations inherent in ToolLLM's evaluation protocol, we propose an alternative evaluation methodology that aligns more closely with real-world scenarios, as illustrated in Figure~\ref{fig:overview}(b). Specifically, we bypass the first evaluation phase of ToolLLM, which assesses the potential of candidate APIs in addressing query $\mathcal{Q}$. Instead, we directly utilize GPT-4 to determine the efficacy of the agent's proposed solution in resolving the query. The pass rate $R$ is thus calculated using the formula:
\begin{equation}
\label{eq:pass_rate}
    R = \frac{\textbf{\#}(\text{Solved})}{\textbf{\#}(\text{Solved}) + \textbf{\#}(\text{Unsolved})}.
\end{equation}
To ensure that all queries in the benchmark, namely ToolBench~\cite{qin2023toolllm}, are solvable using certain APIs from the API pool, we conduct a manual review of all queries. We retain only those queries that can be resolved with specific APIs from this pool. The detailed process is available in Section~\ref{sec:toolbench_filter} of the appendix.

\vspace{-3mm}
\section{AnyTool}
\vspace{-1mm}
Our AnyTool exhibits several distinctive features: Firstly, it eliminates the need
for training external modules, and solely relies on the function calling feature of GPT-4. Secondly, it can directly search the entire API pool, which contains over 16K APIs, using a hierarchical structure and a divide-and-conquer principle. Lastly, it is capable of self-reflection, enabling it to review and analyze unsolved user queries by taking into account reasons for failure and relevant historical contexts.

\vspace{-1mm}
\textbf{Overview.} The overview of AnyTool is depicted in Figure~\ref{fig:method}. It primarily follows a three-step process to efficiently resolve the user query $\mathcal{Q}$. The first step (Section~\ref{sec:api-generator}) involves the creation of an API candidate pool. For efficiency, AnyTool is designed with a hierarchical architecture, taking advantage of the structured API organization available in Rapid API. In the second step (Section~\ref{sec:solver}), a solver attempts to resolve query $\mathcal{Q}$ by utilizing these API candidates. Finally, if the query remains unsolved, AnyTool engages in a self-reflection  process (Section~\ref{sec:self-reflection}) in an attempt to resolve it. A case study is shown in Section~\ref{sec:case_study}.

\vspace{-2mm}
\subsection{API Retriever}
\vspace{-1mm}
\label{sec:api-generator}
\textbf{Structured API Organization in Rapid API.} Rapid API employs a structured system to categorize its extensive collection of 16K+ APIs. Specifically, this organization is divided into three distinct tiers: the first tier is the category level, encompassing various domains such as ``sports'' and ``finance''; the second tier, designated as the tool level, consists of tools that belong to specific categories; and the third tier focuses on individual APIs, with each API belonging to a specific tool, as illustrated in Figure~\ref{fig:method}. This hierarchical arrangement serves as a foundational guideline in the development of our API retriever.

\vspace{-1mm}
\textbf{Hierarchical Structure.} As depicted in Figure~\ref{fig:method}, the structure of our API retriever consists of three tiers. At the initial tier, a meta-agent exists, tasked with dynamically generating a series of category agents in response to the user query $\mathcal{Q}$. The intermediary tier is comprised of multiple category agents, each established by the meta-agent. These agents correspond to individual categories as defined by Rapid API, with their primary objective being to identify the most relevant tools for the query $\mathcal{Q}$ from their respective tool collections. Subsequently, these category agents initiate the creation of various tool agents. It is important to note that each tool agent may manage multiple tools, depending on the decisions made by the category agents. The goal of each tool agent is to search through its managed APIs for those that might solve the query $\mathcal{Q}$, and then add these APIs to an API-candidate pool. Each type of agent possesses its own distinct set of functions. These are illustrated in Figure~\ref{fig:method} and further detailed in Table~\ref{tab:agent_function_list} in the appendix.

\vspace{-1mm}
\textbf{Generation of API-Candidate Pool.} AnyTool is initiated upon receiving a query $\mathcal{Q}$, the function list detailed in Table~\ref{tab:agent_function_list}, and a bootstrap prompt as outlined in Section~\ref{sec:bootstrap_prompt_API_retriever} of the appendix. This process heavily relies on the function calling feature of GPT-4 (refer to Section~\ref{sec:function_calling}). Operating interactively, our system enables agents (starting with the meta-agent) to send requests for calling their managed functions. These functions may involve creating a specific agent (either a category agent or a tool agent) or executing a particular function, in accordance with the historical context.\footnote{Each agent, whether it is a meta-agent, category agent, or tool agent, maintains its own historical context independently.} The requests are parsed, and the corresponding functions are executed. The results produced by these functions are subsequently incorporated into the historical context, which is then returned to the agents. This process repeats continuously until the termination criteria are met. All agents, including meta-agents, category agents, and tool agents, operate independently in a multi-threaded manner, significantly accelerating the process. We maintain a global API candidate pool, allowing each tool agent to add APIs to this pool, using the function ``\texttt{add\_API\_into\_API\_pool}'' (refer to Figure~\ref{fig:method} and Table~\ref{tab:agent_function_list}). All agents cease operations only when a tool agent calls the function ``\texttt{check\_if\_request\_solvable}'' and receives a return value of ``True''. Subsequently, an API-candidate pool is obtained. In addition, we record the historical context and status of each agent. An agent's status is marked as ``Finished'' only if it calls the function ``\texttt{finish\_search}'' during the process. Agents marked as ``Finished'' are excluded in the self-reflection process, which will be described later.

\vspace{-1mm}
\subsection{Solver}
\vspace{-1mm}
\label{sec:solver}
\textbf{Functionality.} The primary goal of the solver is to address the user's query $\mathcal{Q}$, utilizing the generated API candidate pool. It is implemented as a singular agent that leverages the function-calling capabilities inherent in GPT-4. Two potential implementations for the solver are the Depth-First Search-Based Decision Tree (DFSDT) or the Chain of Thought (CoT) approach. A concise overview of the process is provided, with comprehensive details available in ToolLLM~\cite{qin2023toolllm}. The solver activates upon receiving a query $\mathcal{Q}$, in conjunction with a suite of functions, which includes those from the API candidate pool and a distinctive function named ``\texttt{finish}'', as well as a bootstrap prompt detailed in Section~\ref{sec:bootstrap_prompt_solver} of the appendix. The ``\texttt{finish}'' function yields one of three possible outcomes: ``Give Solution'', ``Try Backtrack'', or ``Give Up'', with ``Try Backtrack'' being specific to the DFSDT implementation. Each iteration involves: 1) the solver sending a request to call a function, 2) the interpretation of this request and the execution of the function, and 3) the integration of the function's outcomes into the contextual history, which is then returned to the solver. This cycle continues until the solver gives a ``Give Solution'' or ``Give Up'' decision. Note that when the solver makes a ``Give Up'' decision, it is required to provide both the reason and the function name of the APIs that are irrelevant to the user's query or do not work properly. Self-reflection mechanism is triggered under two scenarios: 1) ``Give Solution'', where GPT-4 reviews the solution and determines that the query remains unresolved, and 2) ``Give Up'', where the solver fails to address the query.

\begin{table*}[!t]
\centering
\caption{Main results on the filtered ToolBench. We use pass rate defined in Eq~\ref{eq:pass_rate} and illustrated in Figure~\ref{fig:overview}(b), as the metric. All results are reproduced. *: OpenAI’s text-embedding-ada-002; Ref.: reference; Avg.: average; SR: self-reflective.}
\vspace{-3mm}
\resizebox{1.0\textwidth}{!}{
\begin{tabular}{l cc ccc cc cc c}
\toprule
 \multirow{2}[3]{*}{Model}& \multirow{2}[3]{*}{API Retriever} & \multirow{2}[3]{*}{Solver} & \multirow{2}[3]{1.3cm}{\centering Use Ref. APIs} &\multicolumn{3}{c}{G1}&\multicolumn{2}{c}{G2}&\multicolumn{1}{c}{G3}&\multirow{2}[3]{*}{Avg. (\%)}\\ 
 \cmidrule(lr){5-7}  \cmidrule(lr){8-9} \cmidrule(lr){10-10}
&  & &  & I (\%) & T (\%) & C (\%) & I (\%) & C (\%) & I (\%)&\\
\midrule
   ToolLLM & OpenAI TE$^*$ & ToolLLaMA w/ DFSDT & &8.7&6.8&12.0&4.7&8.2&10.5&8.5 \\
    ToolLLM & ToolLLM's & ToolLLaMA w/ DFSDT & &28.4&26.3&38.4&21.5&15.1&7.7&22.9\\
    ToolLLM & ToolLLM's & GPT-4 w/ DFSDT &&42.6&46.2&51.4&23.4&24.5&2.6&31.8 \\
    ToolLLM & None & ToolLLaMA w/ DFSDT & \checkmark &29.4&31.8&37.1&19.6&22.4&13.2&25.6\\
    \midrule
    GPT-4& None & GPT-4 w/ CoT & \checkmark &31.3&34.8&47.1&27.1&34.7&2.6&29.6 \\
    GPT-4& None & GPT-4 w/ DFSDT & \checkmark&36.5&49.2&51.4&38.3&39.8&18.4&38.9 \\
    GPT-4 & Plain Agent & GPT-4 w/ DFSDT & &13.9 & 23.5&17.6&13.9&9.2&13.2&15.2\\
    GPT-4 & AutoGen-RAG & GPT-4 w/ DFSDT & &14.8&19.7&19.7 &7.4&9.2&7.9&13.1\\
    \midrule
    GPT-3.5&None &  GPT-3.5 w/ CoT&\checkmark&37.5&37.1&42.9&24.3&22.4&5.3&28.3\\
    GPT-3.5& None & GPT-3.5 w/ DFSDT& \checkmark &39.1&40.2&48.6&31.8&25.5&15.8&33.5\\
\midrule
  AnyTool (Ours) & SR Agent &  SR GPT-4 w/ DFSDT& &52.2&61.4&67.6&58.9&45.9&63.2 &58.2\\
\bottomrule
\end{tabular}
}
\vspace{-3mm}
\label{tab:toolbench}
\end{table*}

\vspace{-2mm}
\subsection{Self-Reflection Mechanism}
\vspace{-1mm}
\label{sec:self-reflection}
If the initial solution fails to resolve user queries, the self-reflection mechanism re-activates AnyTool sequentially, first activating the API retriever and then the solver. It is worth noting that this mechanism can be applied repeatedly until the termination condition is met.

\vspace{-2mm}
\textbf{Self-Reflection in the API Retriever.} Our self-reflection mechanism first identifies the reason why a user query remains unsolved. In instances where the solver opts to ``Give Up'', the rationale provided by the solver is utilized. Conversely, if the solver proposes a solution but GPT-4 assesses that it does not adequately address the query, the reasoning ascribed by GPT-4 is employed. Recall that we maintain a record of historical context for each agent within the API retriever. We initially incorporate the identified reason into all these historical contexts. Owing to the hierarchical design of our API retriever, we systematically re-activate various agents for efficiency purposes, following an ascending order from tool agents, to category agents, and finally to the meta-agent. It is worth noting that only the agents not marked with a ``Finished'' status are re-activated. As a result, this process expands our API-candidate pool, incorporating new APIs that could potentially resolve the user's query.

\vspace{-1.5mm}
\textbf{Self-Reflection in the Solver.} Recall that when the solver makes a ``Give Up'' decision, it is designed to identify the function names of the APIs that are irrelevant to the user's query. For efficiency, we first remove these APIs from the expanded API-candidate pool and exclude items where these APIs are called from the historical context of the solver. The solver is then re-activated with a new bootstrap prompt (refer to Section~\ref{sec:bootstrap_prompt_self_reflection} in the appendix), the updated API-candidate pool, and the cleaned historical context. The remaining process is the same as described in Section~\ref{sec:solver}.

%% file: sections/4-experiments.tex
\vspace{-3mm}
\section{Experiments}
\vspace{-1mm}
\subsection{Setup}
\vspace{-2mm}
\textbf{Benchmarks.} We conduct experiments on two benchmarks: 1) \textit{ToolBench}~\cite{qin2023toolllm}; and 2) our own benchmark, termed \textit{AnyToolBench}. ToolBench comprises six subsets: G1-Instruction (G1-I), G1-Tool (G1-T), G1-Category (G1-C), G2-Instruction (G2-I), G2-Category (G2-C), and G3-Instruction (G3-I). As described at the end of Section~\ref{sec:problem_evaluation}, we perform a manual review on ToolBench to exclude non-solvable queries. Details of this process can be found in Section~\ref{sec:toolbench_filter} of the appendix. After filtering, the remaining queries in these six subsets are 115, 132, 142, 107, 98, and 38, respectively. Unless otherwise specific, we adopt the filtered \textit{ToolBench}. Our benchmark, \textit{AnyToolBench}, includes 400 instances. The process of creating AnyToolBench is detailed in Section~\ref{sec:anytoolbench_construction} of the appendix.

\vspace{-1.5mm}
\textbf{Evaluation Protocol.} We employ the pass rate (as defined in Eq.~\ref{eq:pass_rate}) as our evaluation metric. To assess whether a solution generated by an agent can resolve the query, we use GPT-4-32K. The same prompt utilized in ToolBench is applied when GPT-4 serves as the judge.

\vspace{-1mm}
\textbf{Alignment between GPT-4's Decisions and Decisions Made by Human Evaluators.} We conduct a comparative analysis between decisions made by human evaluators and those generated by GPT-4, focusing on samples from the G1-I subset of ToolBench. Specifically, for each query sample, AnyTool generates a solution, which is then assessed for its feasibility in addressing the query by both human evaluators and GPT-4. Our results reveal that GPT-4's alignment with human evaluation stands at 96.5\%, while that of GPT-3.5 is only 73.9\%. Based on these findings, we exclusively utilize GPT-4 for our evaluations.

\vspace{-2mm}
\subsection{Main Results}
\vspace{-2mm}
We compare our AnyTool with the pioneering ToolLLM~\cite{qin2023toolllm} and its variants, as well as various GPT-4 models tailored for tool utilization.

\vspace{-1.5mm}
\textbf{ToolLLM and Its Variants.} ToolLLM integrates an API retriever\footnote{ToolLLM's API retriever is trained on pair-wise data. Each pair includes a user query and a set of APIs relevant to the query.} and a solver designed to address user queries by employing API candidates produced by the retriever. The solver operates using a finely-tuned LLaMA model, named ToolLLaMA, and employs a depth-first search-based decision tree (DFSDT) algorithm to resolve queries. For each query, ToolBench provides a set of reference APIs that are potentially relevant. These reference APIs offer a means to evaluate the solver's effectiveness by allowing the bypassing of the API retriever step. It is worth noting that additional APIs from the complete API pool, containing over 16,000 APIs, may also contribute to effectively resolving queries. Beyond the original ToolLLM, our experiments also examine two variants: 1) one that substitutes ToolLLaMA with GPT-4 in the solver; 2) another that foregoes the API retriever and relies solely on reference APIs.

\vspace{-1.5mm}
\textbf{Various GPT-4 Models.} The function-calling feature of GPT-4 enables it to use APIs directly for resolving user queries. However, in our setting, we deal with over 16,000 APIs. Integrating all these APIs—each with its unique function description, input, and output—into GPT-4 simultaneously exceeds the maximum context length of the model, even for the version with the largest context length of 128,000 tokens. Therefore, we compare four GPT-4 models: 1) one that uses reference APIs and the Chain of Thought (CoT)~\cite{wei2022chain} algorithm in the solver; 2) another that uses reference APIs and the DFSDT algorithm; 3) a third that employs a plain agent for API retrieval and incorporates the DFSDT algorithm in the solver; 4) a fourth that leverages the Retrieval Augmented Generation (RAG) feature from AutoGen~\cite{augenstein2023factuality} for API retrieval, and uses the DFSDT algorithm to resolve user queries through the selected API candidates.

\vspace{-1mm}
In the implementation of GPT4-plain-agent, we divide the set of over 16K APIs into 33 groups, each containing 500 APIs, with the exception of the 33rd group. These groups are then sequentially processed by GPT-4. The specific task assigned to GPT-4 involves identifying the relevant APIs using the \texttt{add\_API\_into\_API\_pool} function, which integrates them into the API-candidate pool. Refer to Section~\ref{sec:plain_agent} for more details. Information on AutogGen-RAG can be found in Section~\ref{sec:autogen_rag}.

\begin{table}[!t]
\centering
\caption{Main results on our AnyToolBench. All models use DFSDT implementation in the solver. SR: self-reflective; PR: pass rate.}
\vspace{-3mm}
\resizebox{\columnwidth}{!}{
\begin{tabular}{lccc}
\toprule
Method& API Retriever & Solver & PR (\%)\\

\midrule
    ToolLLM & ToolLLM’s & ToolLLaMA  &18.9\\
    ToolLLM & ToolLLM’s & GPT-4 &36.6\\
    GPT-4 & Plain Agent & GPT-4 & 14.0\\
    \midrule
  AnyTool (Ours) & SR Agent & SR GPT-4 &73.8\\
\bottomrule
\end{tabular}
}
\vspace{-4mm}
\label{tab:anytoolbench}
\end{table}

\vspace{-1mm}
\textbf{Main Results on ToolBench.} In Table~\ref{tab:toolbench}, we compare our AnyTool with various ToolLLM variants and GPT-4 models across six subsets of the filtered ToolBench dataset. The results on the original ToolBench are available in Section~\ref{sec:original_toolbench_results}. Both the API retriever and the solver contribute to the final performance. The API retriever's role is to efficiently identify the most pertinent APIs from an extensive collection, while the solver is tasked with generating viable solutions for user queries. Instead of training an API retriever as ToolLLM does, we leverage the powerful function-calling feature of GPT-4 and overcome the challenge posed by its inherent maximum context length limitation, through the implementation of a hierarchical structure. Our self-reflection mechanism applies to both the API retriever and the solver, enabling the whole system to operate in a closed loop. Owing to these factors, our AnyTool significantly outperforms both the original ToolLLM and GPT-4 using reference APIs, by +32.6 and +19.3 points, respectively, in terms of the average pass rate.

\vspace{-1mm}
\textbf{Main Results on AnyToolBench.} AnyToolBench evaluates an agent's capability to resolve user queries utilizing the entire API pool. Consequently, an API retriever is essential in this setting. We do not supply reference APIs for each query; thus, making comparisons with counterparts lacking an API retriever is impractical. In Table~\ref{tab:anytoolbench}, we compare our AnyTool with a top-performing ToolLLM variant and GPT-4, where a plain agent serves as the retriever. The consistent improvements demonstrated by AnyTool over these approaches affirm its effectiveness in a realistic setting.

\begin{table}[!t]
\centering
\small
\caption{Ablation study on the pass rate of main components. ``-'' and ``+'' symbols denote the removal and addition of a component from and into AnyTool, respectively.}
\vspace{-3mm}
\begin{tabular}{lcc}
\toprule
Configuration & G2-I (\%) & G3-I (\%)\\
\midrule
    AnyTool&58.9&63.2\\
    \midrule
    -Hierarchical Structure&22.4&15.8 \\
    -Self-Reflection&19.6&15.8\\
    -DFSDT/+CoT &50.5&60.3\\
\bottomrule
\end{tabular}
\vspace{-2mm}
\label{tab:component}
\end{table}

\vspace{-2.5mm}
\subsection{Ablation Studies}
\vspace{-2mm}
Unless otherwise specific, all ablation studies are conducted on G2-I and G3-I of the filtered ToolBench.

\begin{table}[!t]
\centering
\small
\caption{Ablation study on the pass rate of self-reflection mechanism. All agents include the tool agents, the category agents and the meta-agent.}
\vspace{-3mm}
\begin{tabular}{lcc}
\toprule
Re-Activation & G2-I (\%) & G3-I (\%)\\
\midrule
Tool Agents &43.9&44.7\\
Tool Agents + Category Agents&55.2&55.3\\
All Agents &58.9&63.2\\
\bottomrule
\end{tabular}
\vspace{-4mm}
\label{tab:ablation-Self-Reflection}
\end{table}

\vspace{-1.5mm}
\textbf{Effectiveness of the Main Elements.} Our AnyTool comprises two principal elements: firstly, an API retriever with a hierarchical structure, and secondly, a self-reflection mechanism. In Table~\ref{tab:component}, we examine three distinct configurations of AnyTool. These include: a) substituting our hierarchical API retriever with a flat-structure version, which merges the functions of agents at the category and tool levels (except for ``agent creation'' and ``finish search'' functions) into the function list of the meta-agent; b) eliminating the self-reflection mechanism; and c) substituting the DFSDT algorithm with CoT, thereby disabling the backtracking feature in DFSDT. Our findings demonstrate significant positive effects of both the hierarchical structure and the self-reflection feature on AnyTool's performance. Choosing CoT over DFSDT results in a decline in pass rates by 8.4 and 2.9, respectively.

\vspace{-1.5mm}
\textbf{Self-Reflection Mechanism.} In Section~\ref{sec:self-reflection}, we introduce a self-reflection mechanism that is first applied to the API retriever module. It re-activates various agents in ascending order, from tool agents to category agents, and finally to the meta-agent. In Table~\ref{tab:ablation-Self-Reflection}, we examine the different versions that reactivate distinct types of agents. Reactivating all agents results in the best performance, owing to the larger search space.

\begin{table}[!t]
\centering
\small
\caption{Study on the effects of the API pool's size to the pass rate.}
\vspace{-3mm}
\begin{tabular}{ccc}
\toprule
Size of API Pool & G2-I (\%) & G3-I (\%)\\
\midrule
1,000&18.6&7.9\\
5,000&26.3&23.7\\
10,000&38.1&36.8\\
All&58.9&63.2\\
\bottomrule
\end{tabular}
\vspace{-1.5mm}
\label{tab:pool_size}
\end{table}

\vspace{-1.5mm}
\textbf{Size of the API Pool.} Users typically submit a wide range of queries to the AI system, seeking solutions to real-world problems. To effectively address these queries, the system requires access to a diverse array of APIs. In general, a larger API pool is more likely to successfully resolve user queries, as it offers a higher probability of containing relevant APIs. This hypothesis is evaluated by randomly selecting subsets of APIs from the complete pool and using only these subsets to address user queries with our AnyTool. The results, presented in Table~\ref{tab:pool_size}, support our hypothesis.

\begin{table}[!t]
\centering
\small
\caption{Study on the maximal size of API-candidate pool.}
\vspace{-3mm}
\resizebox{\columnwidth}{!}{
\begin{tabular}{ccc}
\toprule
Maximal Size of API-Candidate Pool & G2-I (\%) & G3-I (\%)\\
\midrule
16&49.5&42.1\\
32&58.9&55.3\\
64&58.9&63.2\\
\bottomrule
\end{tabular}
}
\vspace{-3mm}
\label{tab:size-candidate-pool}
\end{table}

\vspace{-1.5mm}
\textbf{Maximal Size of the API-Candidate Pool.} AnyTool operates through a two-step process—the solver addresses queries by using an API-candidate pool, which is generated by our hierarchical API Retriever. One termination criterion for the API retriever is the fullness of this pool. We examine the impact of the maximal size of the API-candidate pool as shown in Table~\ref{tab:size-candidate-pool}. We observe that a pool size of 64 nearly reaches saturation in terms of performance.

\begin{table}[!t]
\centering
\small
\caption{We study the maximum number of tools that a tool agent can manage in our API retriever.}
\vspace{-3mm}
\begin{tabular}{ccc}
\toprule
 Maximum Number of Tools & G2-I (\%) & G3-I (\%)\\
\midrule
3&48.6&42.1\\
5&58.9&57.9\\
10&52.3&39.5\\
\bottomrule
\end{tabular}
\vspace{-3mm}
\label{tab:maximum-number-tools}
\end{table}

\vspace{-1.5mm}
\textbf{Tool Agent in API retriever.} Our API retriever is designed with a hierarchical structure, in which the tool agents at the bottom layer directly add APIs that may potentially address user queries, into the API-candidate pool. As described in Section~\ref{sec:api-generator}, a tool agent can manage a maximum of $K$ tools existing in Rapid API. We examine the value of $K$ in Table~\ref{tab:maximum-number-tools}. A trade-off is observed: managing too many tools (e.g., $K=10$) leads to a larger search space and may cause overlooking of relevant APIs, while managing too few tools (e.g., $K=3$) might result in lower recall.

\begin{figure}[!t]
    \centering
\includegraphics[width=0.99\linewidth]{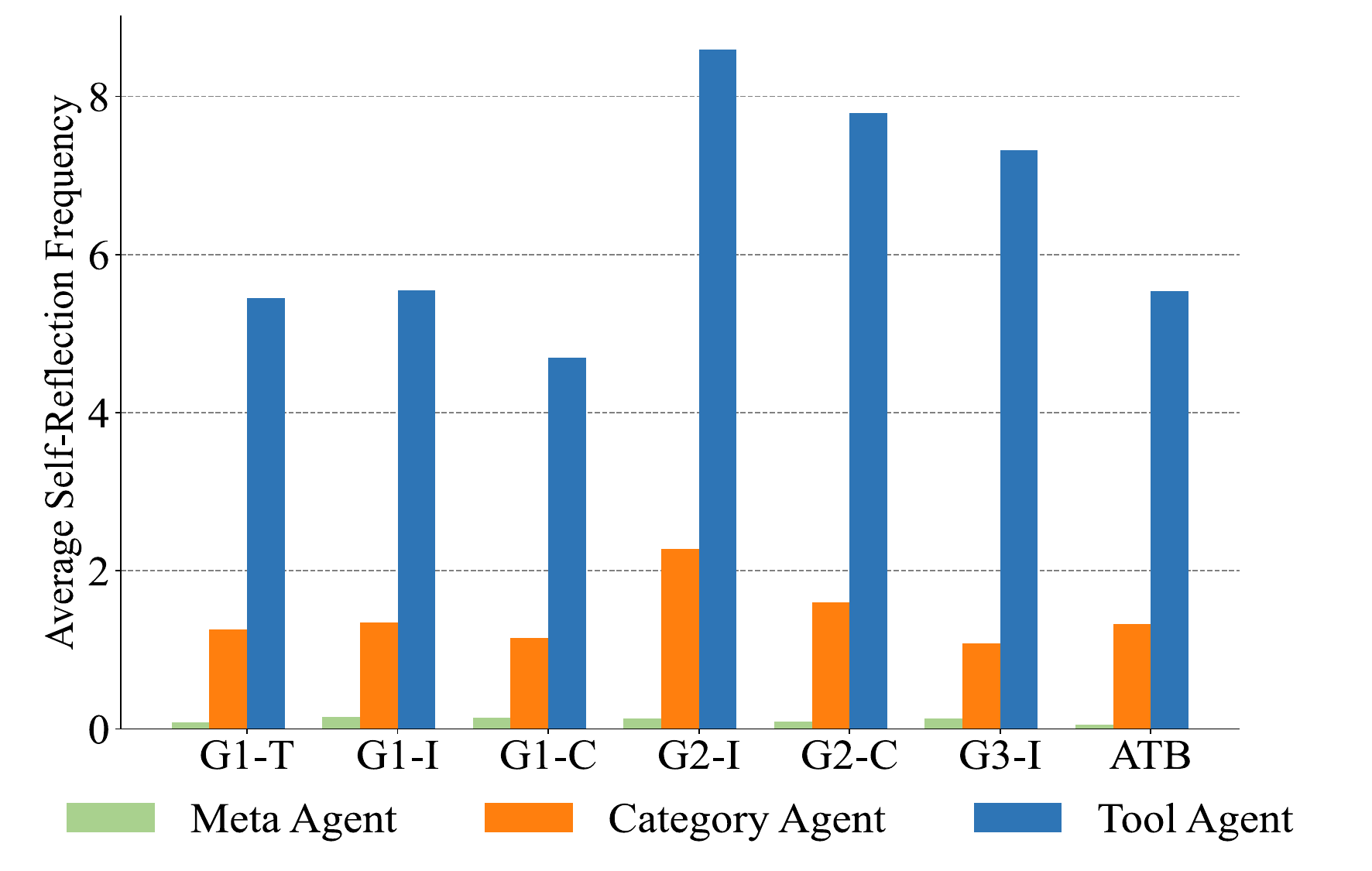}
\vspace{-4mm}
    \caption{Statistics of average self-reflection frequency. ATB: AnyToolBench.}
    \vspace{-4mm}
    \label{fig:self_reflection_frequency}
\end{figure}

\vspace{-1.5mm}
\textbf{Statistics of Self-Reflection Frequency.} In Figure~\ref{fig:self_reflection_frequency}, we report the average self-reflection frequency across all instances within each subset of the filtered ToolBench and our AnyToolBench. As described in Section~\ref{sec:self-reflection}, we reactivate various agents in ascending order. Consequently, the frequency of tool agents is much higher than that of category agents and meta-agent. Additionally, calculating the processing time for resolving queries with AnyTool is infeasible. AnyTool relies on the function-calling feature of GPT-4, whose server response is often unstable.

\begin{figure}[!t]
    \centering
\includegraphics[width=0.99\linewidth]{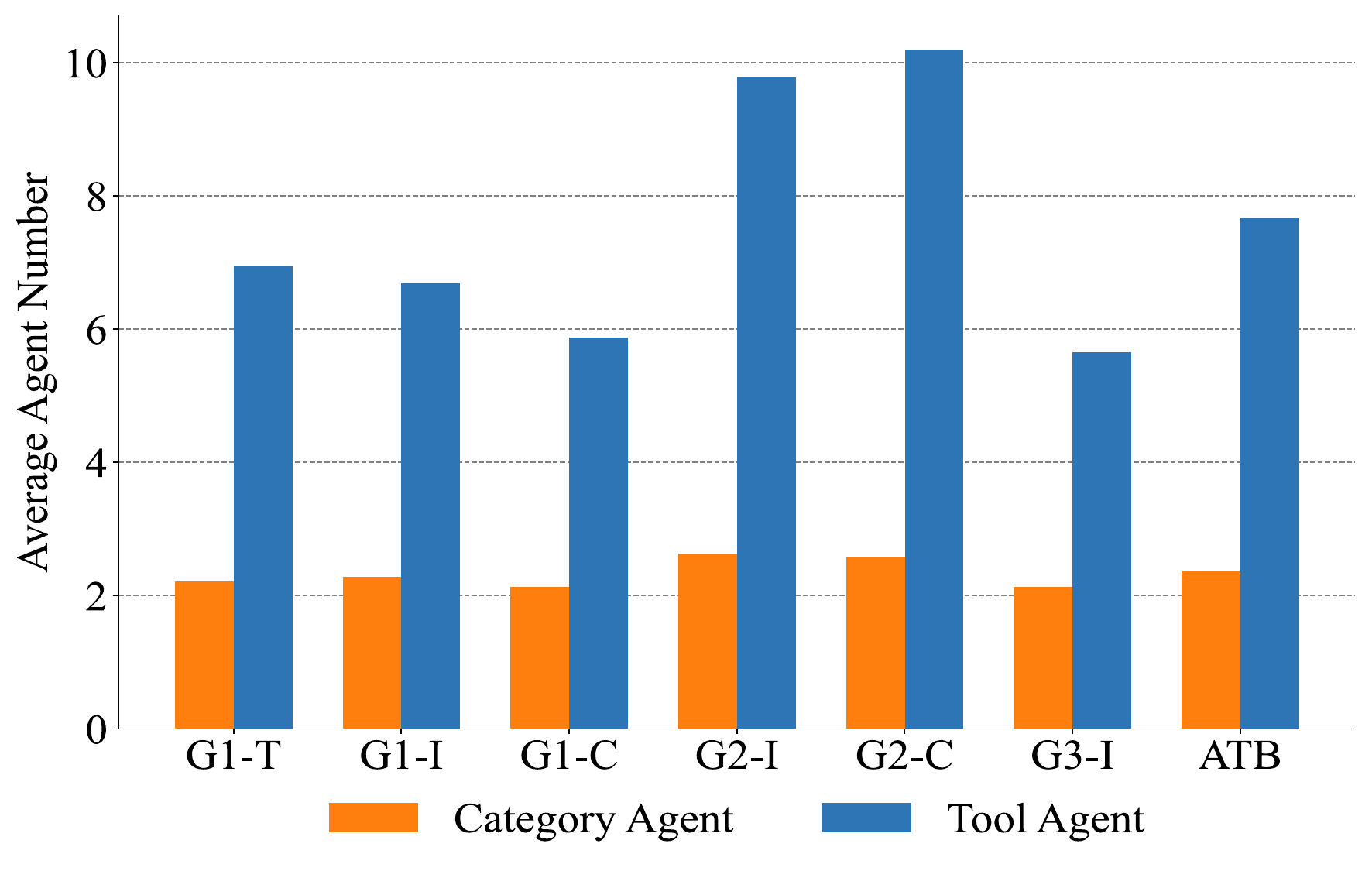}
\vspace{-4mm}
    \caption{Statistics of average agent quantity.}
    \vspace{-5mm}
    \label{fig:agent_number}
\end{figure}

\vspace{-1.5mm}
\textbf{Agent Quantity in API Retriever.} 
The API retriever of AnyTool is hierarchically structured. Depending on the nature of user queries, the meta-agent can dynamically create a varying number of category agents. This process is analogous to the way category agents create tool agents. The average number of agents across all instances in each subset of the filtered ToolBench and our AnyToolBench is depicted in Figure~\ref{fig:agent_number}.

%% file: sections/5-conclusion.tex
\vspace{-3mm}
\section{Conclusion}
\vspace{-1mm}
In this work, we introduce AnyTool, an advanced agent capable of harnessing 16K+ APIs to effectively handle realistic user inquiries. The core of AnyTool is a hierarchical API retriever coupled with a solver. Additionally, it incorporates a unique self-reflection mechanism, enhancing its proficiency in responding to user queries. We also revise the prior evaluation protocol to better reflect real-world application scenarios. Rigorous experiments conducted on ToolBench and our AnyToolBench demonstrate our approach's superiority over established models. Finally, we highlight two future research directions: 1) optimizing the organization of APIs for improved performance and efficiency; 2) developing an advanced open-source LLM specifically for API utilization, which could facilitate local deployments.

%% file: sections/7-impact.tex
\section*{Impact Statements}
Although AnyTool significantly enhances the effectiveness of resolving user queries through various tools, its performance in extremely complex scenarios has not been verified, owing to the absence of appropriate datasets. Furthermore, as AnyTool relies on the function-calling feature of GPT-4, the capabilities of GPT-4 also affect the feasibility of the solutions it generates.

%% file: sections/6-appendix.tex
\newpage
\appendix
\onecolumn
\section{More Implementation Details and Experimental Results}

\begin{table*}[!t]
    \centering
    \small
    \caption{Function list of each type of agent. $^*$: descriptions of input, output and functionality.}
    \vspace{-3mm}
    \begin{tabular}{l l p{4cm} l l}
            \toprule
Type & Function Name & Functionality & Input & Output\\
             \midrule
\multirow{4}{*}{Meta Agent}& \texttt{create\_agent\_category\_level} &  Create a category agent. & Category name & Category agent\\
& \texttt{get\_tools\_in\_category} & Get tool names under a category. & Category name & [Tool names]\\
& \texttt{get\_tool\_descriptions} & Get description of each tool. & [Tools] & [Tool descriptions]\\
& \texttt{finish\_search} & Send out finish signal. & None & None\\
\midrule
\multirow{4}{*}{Category Agent}& \texttt{create\_agent\_tool\_level} & Create a tool agent. & [Tools] & Tool agent\\
& \texttt{get\_tools\_in\_category} & Get tool names under a category. & Category name & [Tool names]\\
& \texttt{get\_tool\_descriptions} & Get description of each tool. & [Tools] & [Tool descriptions]\\
& \texttt{finish\_search} & Send out finish signal.& None & None \\
\midrule
\multirow{7}{*}{Tool Agent}& \texttt{add\_API\_into\_API\_pool} & Add APIs into candidate pool.& [APIs] & None \\
&  \texttt{get\_APIs\_in\_tool}& Get API names under a tool. & Tool name & [API names]\\
&  \texttt{get\_API\_detail}& Get detail$^*$ of each API. & [API names] & [API details]\\
&  \multirow{3}{*}{\texttt{check\_if\_request\_solvable}}& Check whether the query is solvable using the current candidate pool. & \multirow{3}{*}{None} & \multirow{3}{*}{True\textbackslash False} \\
&  \texttt{finish\_search}& Send out finish signal.& None & None \\
            \bottomrule        
        \end{tabular}
     \vspace{-2mm}
    \label{tab:agent_function_list}
\end{table*}

\subsection{More Implementation Details of AnyTool}
\label{sec:details_anytool}
For the solver implementing DFSDT, we set the maximum number of API calls to 10. Additionally, for our AnyTool, we establish a limit of 200,000 tokens for efficiency. This limit encompasses the token consumption by various components, including the meta-agent, the tool agents, the category agents, the solver, and the self-reflection mechanism.

\subsection{Detailed Function List}
\label{sec:details_function_list}
We provide the function list of each type of agent in Table~\ref{tab:agent_function_list}.

\subsection{Results on the Original ToolBench}
\label{sec:original_toolbench_results}
We also provide the results on the original ToolBench~\cite{qin2023toolllm} without undergoing filtering process. In the original ToolBench, each subset comprises 200 queries, except for G3-I, which contains 100 queries. Note that the original ToolBench includes non-solvable queries. We test all queries, regardless of whether they are solvable or not, using pass rate defined in Eq~\ref{eq:pass_rate} and illustrated in Figure~\ref{fig:overview}(b), as the metric. All results are reproduced. As shown in Table~\ref{tab:ori_toolbench}, our AnyTool outperforms all ToolLLM~\cite{qin2023toolllm} variants.

\begin{table}[t]
\centering
\caption{Results on the original ToolBench~\cite{qin2023toolllm}. Note that the original ToolBench includes non-solvable queries. We use pass rate defined in Eq~\ref{eq:pass_rate} and illustrated in Figure~\ref{fig:overview}(b), as the metric. All results are reproduced. Ref.: reference; Avg.: average; SR: self-reflective.}
 \vspace{-3mm}
\resizebox{1.0\textwidth}{!}{\begin{tabular}{l cc ccc cc cc c}
\toprule
 \multirow{2}[3]{*}{Model}& \multirow{2}[3]{*}{API Retriever} & \multirow{2}[3]{*}{Solver} & \multirow{2}[3]{1.3cm}{\centering Use Ref. APIs} &\multicolumn{3}{c}{G1}&\multicolumn{2}{c}{G2}&\multicolumn{1}{c}{G3}&\multirow{2}[3]{*}{Avg. (\%)}\\ 
 \cmidrule(lr){5-7}  \cmidrule(lr){8-9} \cmidrule(lr){10-10}
&  & &  & I (\%) & T (\%) & C (\%) & I (\%) & C (\%) & I (\%)&\\
\midrule
    ToolLLM & ToolLLM's & ToolLLaMA w/ DFSDT &&24.0&23.0&37.5&17.5&16.5&4.0&20.4\\
    ToolLLM &ToolLLM's & GPT-4 w/ DFSDT &&32.0&43.5&46.5&30.0&33.0&8.0&32.2\\
  AnyTool (Ours) & SR Agent &  SR GPT-4 w/ DFSDT& &46.0&54.0&53.0&37.0&46.5&32.0&44.8\\
\bottomrule
\end{tabular}}
\label{tab:ori_toolbench}
\end{table}

\subsection{GPT-4 with Various Plain Agents}
\label{sec:plain_agent}

In Table~\ref{tab:toolbench} of the main paper, we present a comparison between our AnyTool and a GPT-4 variant. This variant employs a plain agent as the API retriever, which is limited to accessing only the \textit{names} of tools and APIs. It utilizes the \texttt{add\_API\_into\_API\_pool} function to incorporate APIs into the API candidate pool. When an API is added to the pool, we use the \texttt{check\_if\_request\_solvable} function to determine whether the current API candidates are adequate for addressing the query. If the evaluation returns ``True'', the solver begins to resolve the query using the API candidates with the DFSDT algorithm. Note that the plain agent does not involve any self-reflection mechanism.

In Table~\ref{tab:ablation_plain}, we explore alternative configurations where the plain agent could access both \textit{names} and \textit{detailed descriptions} of tools and APIs (every 100 APIs a group), or even comprehensive information including the \textit{names}, \textit{descriptions}, and specific \textit{API details} (every 50 APIs a group). Our findings suggest that the addition of more detailed information leads to only marginal improvements in performance. In contrast, our AnyTool exhibits superior performance, which can be attributed to its hierarchical structure.

\begin{table}[!t]
\centering
\caption{Comparison of AnyTool and GPT-4 using various plain agents as the API retriever. The only difference among these plain agents lies in the information they can access.}
\vspace{-2mm}
\begin{tabular}{lcc}
\toprule

GPT-4 Variant& G2-I (\%) & G3-I (\%)\\
\midrule
    w/ Names&13.1&13.2\\
     w/ Names+Description &15.9 &13.2  \\
w/ Names+Description+Details&13.1  & 13.2\\
\midrule
AnyTool (Ours) &58.9 & 63.2\\
\bottomrule
\end{tabular}

\vspace{-2mm}
\label{tab:ablation_plain}
\end{table}
\subsection{GPT-4 with Various AutoGen-RAG Agents}
\label{sec:autogen_rag}

Retrieval-augmented generation (RAG) operates by receiving an input and sourcing a collection of pertinent or corroborative documents from a reference, such as Wikipedia. These documents are then combined with the initial input prompt to provide context. This enriched input is subsequently processed by LLMs to generate the final output. The RAG method enhances the performance of LLMs in situations that require accurate factual information.

In Table~\ref{tab:toolbench} of the main paper, we present a version of GPT-4 designed for tool utilization. This version employs AutoGen-RAG as the API retriever. The embedding model, known as ``all-mpnet-base-v2''\footnote{\url{https://huggingface.co/sentence-transformers/all-mpnet-base-v2}}, is utilized in this version. Specifically, we integrate the category names, tool names, API names, and their descriptions into a document, which is then divided into numerous text segments, each containing 1,000 tokens. Then, given a user query, AutoGen-RAG identifies the most relevant segments based on the embedding similarities between the user query and each text segment. Finally, we use GPT-4 to extract the most relevant API candidates from the selected text segments.

We provide another variant, where OpenAI's ``text-embedding-ada-002'' is used as the embedding model. The comparison with our AnyTool is shown in Table~\ref{tab:ablation_rag}.

\begin{table}[!t]
\centering
\caption{Comparison of AnyTool and GPT-4 using various AutoGen-RAG agents as the API retriever. The only difference among
these AutoGen-RAG agents lies in the embedding model they use.}
\vspace{-2mm}
\begin{tabular}{lcc}
\toprule
Embedding Model& G2-I (\%) & G3-I (\%)\\
\midrule
text-embedding-ada-002&8.4&7.9\\
all-mpnet-base-v2 &7.4&7.9\\
\midrule
AnyTool (Ours) & 58.9&63.2 \\
\bottomrule
\end{tabular}

\vspace{-2mm}
\label{tab:ablation_rag}
\end{table}

\subsection{Consumption Analysis}
In our analysis of resource consumption by AnyTool for solving queries across all datasets, we find that, on average, each query consumes $13.5\times 10^4$ tokens, identifies 14.1 API candidates, and involves 43.3 OpenAI API calls and 4.6 self-reflections. Table~\ref{tab:cnsumption_analysis} presents the statistics for each dataset. Additionally, calculating the processing time for resolving queries with AnyTool is infeasible. AnyTool relies on the function-calling feature of GPT-4, whose server response is often unstable.

\begin{table}[t]
\vskip 0.15in
\begin{center}
\caption{Consumption statistics for each dataset.}
\vspace{-3mm}
\begin{tabular}{lcccccccc}
\toprule
\multirow{2}[3]{*}{Statistics}&\multicolumn{3}{c}{G1}&\multicolumn{2}{c}{G2}&\multicolumn{1}{c}{G3}&\multirow{2}[3]{*}{ATB}&\multirow{2}[3]{*}{Avg. }\\ 
 \cmidrule(lr){2-4}  \cmidrule(lr){5-6} \cmidrule(lr){7-7}
& I  & T  & C  & I  & C  & I &&\\
\midrule
   Average Token Consumption ($\times10^4$)&13.6&12.1&8.5&17.7&14.8&16.2&12.2&13.6\\
   Average Call Number &39.3&38.8&33.8&54.0&57.6&35.7&44.2&43.3\\
   Average Self-Reflection Number &4.2&3.8&4.1&5.7&5.2&5.1&4.0&4.6\\
   Average API Candidate Number &13.8&13.0&7.7&16.8&16.0&16.3&14.9&14.1\\
\bottomrule
\end{tabular}
\label{tab:cnsumption_analysis}
\end{center}
\vskip -0.1in
\end{table}

\subsection{Filtering Process for ToolBench}
\label{sec:toolbench_filter}
We primarily screen out non-solvable queries in ToolBench based on the following principles:
\begin{itemize}
    \item Queries lacking essential information, such as unspecified phone numbers or ambiguous references like ``my friend''. These are inherently non-solvable since APIs require explicit input parameters.
    \item Queries containing fake parameters, such as non-existent URLs.
    \item Queries that specify a specific API are filtered out because they do not represent realistic scenarios. Moreover, if the problem can be solved using another API, it is difficult to determine whether it counts as a resolution.
    \item Unreasonable queries, such as asking for information about popular movies on YTS, which are too broad in scope and difficult to evaluate.
\end{itemize}

\begin{table}[!t]
    \centering
    \caption{Examples of our AnyToolBench.}
     \vspace{-3mm}
        \begin{tabular}{p{16.6cm}}
            \toprule
            I am creating an art project about the influence of music on visual arts and for my centerpiece, I would love to have an AI-generated image based on the current number one hit song on the Billboard Hot 100 chart. Could you provide me with such an image that encapsulates the essence of the song 'Bad Habit' by Steve Lacy? \\
            \midrule
            For a business presentation on global trends in music and sports performance analysis, could you provide the top streaming songs on Spotify for the most recent available global chart data, along with the corresponding 'hello world' placeholder text that will be used for introducing programmatic greetings, and the win-loss records for NFL teams from the 2022 season to illustrate the competitive landscape?\\
            \midrule
            Could you analyze potential profit or loss from bitcoin arbitrage among exchanges, considering the market order fees, and check if the IP 23.129.64.215 is flagged for any suspicious activity, and why? I'm interested in arbitrage between Bitfinex, Kraken, and Bittrex for BTC/USD and knowing what risks I might face using the mentioned IP address for transactions.\\
            \midrule
            I plan to improve my daily fitness level, but I always lack proper planning. My current weight is 70 kilograms and my height is 1.75 meters. Given this, could you provide me a health plan regarding the weather condition for outdoor activities in New York for the next five days and the nutrition I intake by usually eating salad?\\
            \bottomrule        
        \end{tabular}
    \vspace{-3mm}
    \label{tab:examples_anytoolbench}
\end{table}

\subsection{Construction of AnyToolBench}
\label{sec:anytoolbench_construction}
We provide GPT-4 with several functions to freely explore the entire API pool, including \{\texttt{get\_tools\_in\_category}, 
\texttt{get\_tool\_descriptions}, \texttt{get\_APIs\_in\_tool}, \texttt{get\_API\_detail}\}. The functionality of these functions are listed in Table~\ref{tab:agent_function_list}.
GPT-4 then utilizes the \texttt{add\_API\_into\_API\_pool} function to incorporate the selected APIs into an API candidate pool. Following this step, GPT-4 generates the required parameters for these APIs and formulates queries based on the actual responses from these APIs. We also prompt GPT-4 to generate a solution for each query, which significantly reduces the potential for hallucinations—the queries may be formulated without utilizing the APIs. Moreover, we enhance the quality of these queries by verifying that the provided reference solutions truly resolve the queries. This rigorous process ensures that every query in our dataset is solvable. The prompt for constructing AnyToolBench is detailed in Section~\ref{sec:prompt_atb}. We show some examples of our AnyToolBench in Table~\ref{tab:examples_anytoolbench}.

\section{Prompts}
\subsection{Bootstrap Prompt for the API Retriever}
\label{sec:bootstrap_prompt_API_retriever}
The API retriever is composed of a meta-agent along with several category agents and tool agents. The bootstrap prompts for these three types of agents are presented in Table \ref{tab:prompt_meta_agent}, Table \ref{tab:prompt_category_agent}, and Table \ref{tab:prompt_tool_agent}, respectively.

\begin{table}[!t]
    \centering
    \caption{Bootstrap prompt for meta-agent.}
     \vspace{-3mm}
        \begin{tabular}{p{16.6cm}}
            \toprule
            {\texttt{You are APIGPT, with access to a database of APIs. This database is organized into the following categories: \{categories\}. Your task is to help users identify the relevant categories for their needs. To do this, you can use the 'query\_tools\_in\_category' function to retrieve the available tools within a specific category. If you are unsure about the functionality of some tools, the 'get\_tools\_descriptions' function can be used to obtain detailed information about these tools. This information will aid you in understanding the general functionality of each category. Additionally, the 'create\_agent\_category\_level' function allows you to assign a relevant category to an agent, with each agent being assigned only one category. However, you can assign multiple categories to different agents. It is important to explore as many categories as possible, as the solution to a query may be found in unexpected categories. Remember, your goal is not to answer the query directly but to identify all potentially relevant categories and assign them to agents. Once you have completed the assignment, call the 'Finish' function. At each step, you should briefly analyze the current status and determine your next action, including the function calls needed to execute your step. Keep your analysis concise, ideally no longer than three sentences.}}\\
            \bottomrule        
        \end{tabular}
    \vspace{-3mm}
    \label{tab:prompt_meta_agent}
\end{table}

\begin{table}[!t]
    \centering
    \caption{Bootstrap prompt for category agent.}
     \vspace{-3mm}
        \begin{tabular}{p{16.6cm}}
            \toprule
            {\texttt{You are APIGPT, with access to a database of APIs categorized into various groups. Each category contains numerous tools, and each tool encompasses multiple APIs. Your task is to assist users in finding relevant tools within a specific category. If uncertain about the functionality of some tools, use the 'get\_tools\_descriptions' function to obtain detailed information. Then, employ the 'create\_agent\_tool\_level' function to allocate a subset of pertinent tools to an agent, ensuring that similar tools are assigned to the same agent and limiting the allocation to no more than five tools per agent. You may assign different subsets to multiple agents. Remember, your role is not to answer queries directly, but to assign all possible tools. Once you complete the assignment, or if you determine the query is irrelevant to the tools in the specified category, invoke the 'Finish' function. Execute each step by calling the appropriate functions, and keep your thought process concise, ideally within three sentences.}}\\
            \bottomrule        
        \end{tabular}
    \vspace{-3mm}
    \label{tab:prompt_category_agent}
\end{table}

\begin{table}[!t]
    \centering
    \caption{Bootstrap prompt for tool agent.}
     \vspace{-3mm}
        \begin{tabular}{p{16.6cm}}
            \toprule
            {\texttt{You are APIGPT with access to a database of APIs, categorized into various sections. Each category contains multiple tools, and each tool encompasses numerous APIs. Your task is to assist users in finding relevant APIs within the tools '\{tools\}' of the '\{category\}' category. You will be provided with descriptions and details of these tools and their APIs. Upon identifying relevant API names, use the 'add\_apis\_into\_api\_pool' function to add them to the final API list. If you conclude that all possible APIs have been explored, or if there are no relevant APIs in these tools, invoke the Finish function. During the process, you may receive feedback on these APIs. At each step, ensure to execute your actions using the appropriate functions. Keep your responses concise, ideally within three sentences.}}\\
            \bottomrule        
        \end{tabular}
    \vspace{-3mm}
    \label{tab:prompt_tool_agent}
\end{table}

    \subsection{Bootstrap Prompt for the Solver}
\label{sec:bootstrap_prompt_solver}
We adapt the prompt from ToolLLM~\cite{qin2023toolllm} to include a ``give\_up'' option without restarting. Furthermore, we prompt it to provide a reason when choosing either ``give\_up\_and\_restart'' or ``give\_up''. The reason should mention specific function names. Table~\ref{tab:prompt_solver} details the prompt for the DFSDT implementation. The task description includes descriptions of accessible functions; therefore, it should be updated to reflect changes in the API candidate pool.

\begin{table}[!t]
    \centering
    \caption{Bootstrap prompt for the solver.}
     \vspace{-3mm}
        \begin{tabular}{p{16.6cm}}
            \toprule
            {\texttt{You are AutoGPT, you can use many tools (functions) to do the following task. First I will give you the task description, and your task start. At each step, you need to give your thought to analyze the status now and what to do next, with a function call to actually excute your step. After the call, you will get the call result, and you are now in a new state. Then you will analyze your status now, then decide what to do next... After many (Thought-call) pairs, you finally perform the task, then you can give your finial answer. If you feel you cannot solve the task or can only solve it partially, you should choose to give up and give your reason which should mention the names of the failed functions. Remember: 1.the state change is irreversible, you can't go back to one of the former state, if you want to restart the task, say "I give up and restart" and give the reason. 2.All the thought is short, at most in 5 sentence. 3.You can do more then one try, so if your plan is to continuously try some conditions, you can do one of the conditions per try. Let's Begin! Task description: \{task\_description\} }}\\
            \bottomrule        
        \end{tabular}
    \label{tab:prompt_solver}
\end{table}

\vspace{-1mm}
\subsection{Bootstrap Prompt for the Self-Reflection Mechanism}
\label{sec:bootstrap_prompt_self_reflection}
Self-reflection mechanism re-activates AnyTool sequentially, first activating the API retriever and then the solver. Owing to the hierarchical design
of our API retriever, we systematically re-activate various agents, following an ascending order
from tool agents, to category agents, and finally to the meta-agent. The prompts for re-activating the tool agents, the category agents and the meta-agent are presented in Table~\ref{tab:re-activating_tool_agents}, Table~\ref{tab:re-activating_category_agents}, and Table~\ref{tab:re-activating_meta_agent}, respectively.

\begin{table}[!t]
    \centering
    \caption{Bootstrap prompt for re-activating tool agents.}
     \vspace{-3mm}
        \begin{tabular}{p{16.6cm}}
            \toprule
            {\texttt{The current APIs have failed to solve the query, resulting in: \{fail\_reason\}. You need to analyze this result and seek additional APIs. It's possible that the tools lack the relevant APIs. In such cases, you should call the Finish function. Remember not to invent tool or API names.}}\\
            \bottomrule        
        \end{tabular}
    \label{tab:re-activating_tool_agents}
\end{table}

\begin{table}[!t]
    \centering
    \caption{Bootstrap prompt for re-activating category agents.}
     \vspace{-3mm}
        \begin{tabular}{p{16.6cm}}
            \toprule
            {\texttt{The current APIs have failed to solve the query, and the reason is: \{fail\_reason\}. Please consider assigning more unexplored tools to the agents.}}\\
            \bottomrule        
        \end{tabular}
    \label{tab:re-activating_category_agents}
\end{table}

\begin{table}[!t]
    \centering
    \caption{Bootstrap prompt for re-activating meta-agent.}
     \vspace{-3mm}
        \begin{tabular}{p{16.6cm}}
            \toprule
            {\texttt{The current APIs have failed to solve the query, and the reason is: \{fail\_reason\}. Please consider assigning more unexplored categories to the agents.}}\\
            \bottomrule        
        \end{tabular}
    \vspace{-1mm}
    \label{tab:re-activating_meta_agent}
\end{table}

\vspace{-1mm}
\subsection{Prompt for Creating AnyToolBench}
\label{sec:prompt_atb}
This can be found in Table~\ref{tab:create_anytoolbench}.

\begin{table}[!t]
    \centering
    \caption{Prompt for Creating AnyToolBench.}
     \vspace{-3mm}
        \begin{tabular}{p{16.6cm}}
            \toprule
{\texttt{Your task is to interact with a sophisticated database of tools and functions, often referred to as APIs, to construct a user query that will be answered using the capabilities of these APIs. This database is organized into various categories, indicated by \{categories\}. To guide your exploration and selection of the appropriate APIs, the database offers several meta functions:}

\texttt{Exploration Functions:}

\texttt{1. Use get\_tools\_in\_category to explore tools in a specific category.}

\texttt{2. Employ get\_apis\_in\_tool to discover the list of APIs available within a selected tool.}

\texttt{3. If you need detailed information about a tool, get\_tool\_descriptions will provide it.}

\texttt{4. For in-depth understanding of an API's functionality, turn to get\_api\_details.}

\texttt{Selection and Testing Functions:}

\texttt{1. As you identify relevant functions, add them to your working list using add\_apis\_into\_api\_pool.}

\texttt{2. Test these functions by synthesizing and applying various parameters. This step is crucial to understand how these functions can be practically applied in formulating your query.}

\texttt{3. Should you find any function obsolete or not fitting your query context, remove them using remove\_apis\_from\_api\_pool.}

\texttt{Query Formulation Guidelines:}

\texttt{1.Your formulated query should be comprehensive, integrating APIs from 2 to 5 different categories. This cross-functional approach is essential to demonstrate the versatility and broad applicability of the database.}

\texttt{2.Avoid using ambiguous terms. Instead, provide detailed, specific information. For instance, if your query involves personal contact details, use provided placeholders like \{email\} for email, \{phone number\} for phone number, and URLs like \{url\} for a company website.}

\texttt{3.The query should be relatable and understandable to users without requiring knowledge of the specific tools or API names used in the background. It should reflect a real-world user scenario.}

\texttt{4. Aim for a query length of at least thirty words to ensure depth and complexity.}

\texttt{Final Steps:}

\texttt{1.Once you've crafted the query, use the Finish function to submit it along with the corresponding answer. The answer should be direct and concise, addressing the query without delving into the operational plan of the APIs.}

\texttt{2.Remember, the total number of calls to the initial meta functions should not exceed 20.}

\texttt{3.Consider various use cases while formulating your query, such as data analysis in business contexts or educational content in academic settings. Your approach should be creative and inclusive, catering to users with different skill levels and cultural backgrounds. Ensure that the query is globally relevant and straightforward, serving a singular purpose without diverging into unrelated areas. The complexity of your query should stem from the synthesis of information from multiple APIs.}}\\
            \bottomrule        
        \end{tabular}
    \label{tab:create_anytoolbench}
\end{table}

\vspace{-2mm}
\section{Case Study}
\label{sec:case_study}
\begin{figure}[!t]
    \centering
\includegraphics[width=0.98\linewidth]{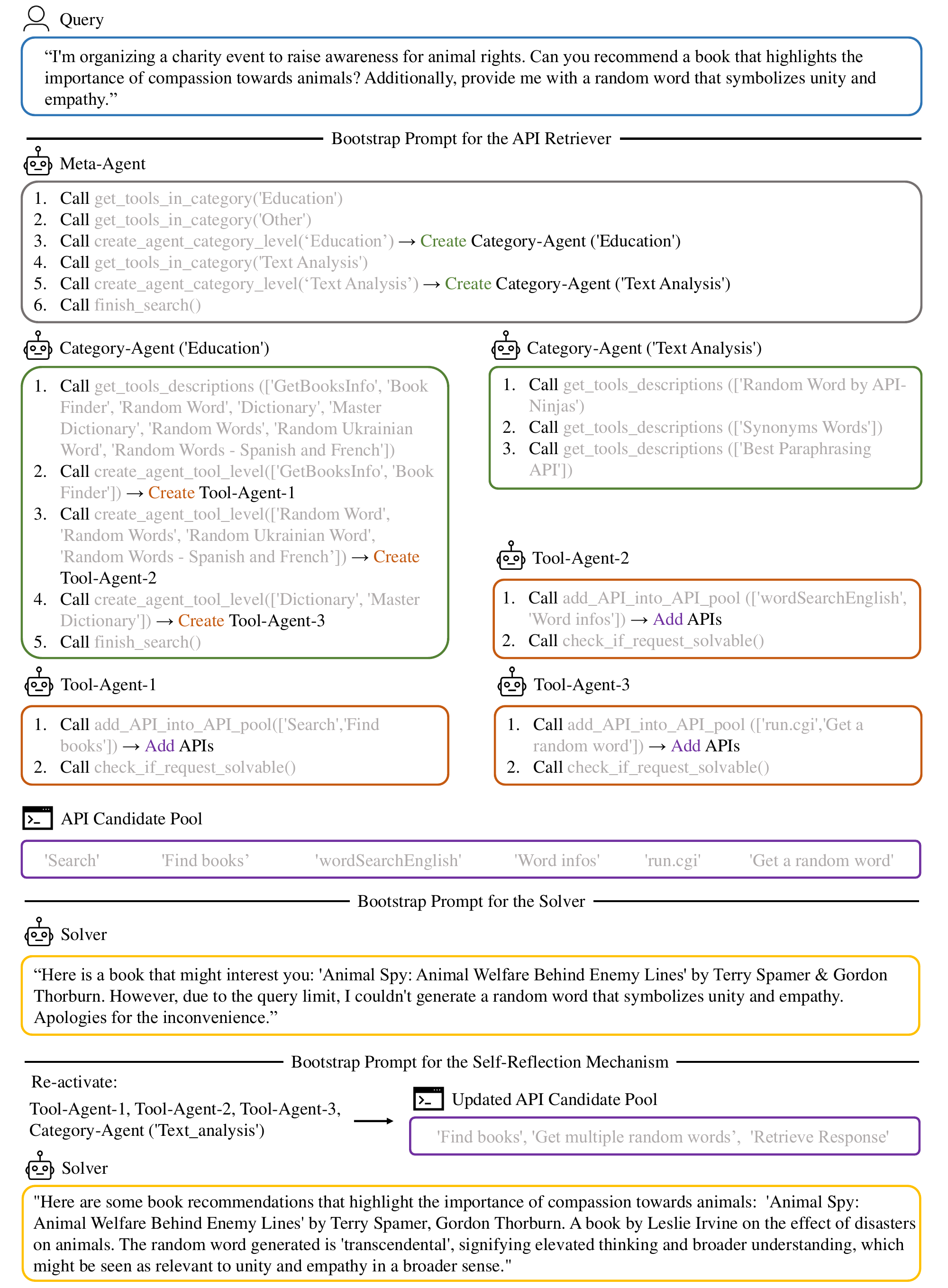}
\vspace{-2mm}
    \caption{Illustration of a case study.}
    \label{fig:case_study}
\end{figure}

In Figure~\ref{fig:case_study}, we present a case study that demonstrates the process of resolving a user query using AnyTool. The self-reflection mechanism reactivates the tool, category, and the meta agents sequentially. It is worth noting that not all agents are reactivated. Subsequently, the solver is reactivated to attempt addressing the user query again, utilizing the updated API candidate pool. This self-reflection mechanism can be employed multiple times until the termination criteria are met—either the query is regarded as solved by the evaluator, or the number of self-reflections reaches the maximum limit.

